\def\year{2022}\relax
\documentclass[letterpaper]{article} 

\usepackage{aaai22}  
\usepackage{times}  
\usepackage{helvet}  
\usepackage{courier}  
\usepackage[hyphens]{url}  
\usepackage{graphicx} 
\usepackage{natbib}  
\usepackage{caption} 
\DeclareCaptionStyle{ruled}{labelfont=normalfont,labelsep=colon,strut=off} 
\usepackage{algorithm}
\usepackage{algorithmic}
\usepackage{subcaption}
 \usepackage{amsmath}
\usepackage{newfloat}
\usepackage{listings}
\floatstyle{ruled}
\newfloat{listing}{tb}{lst}{}
\floatname{listing}{Listing}
\graphicspath{ {./images/} }
\title{Robust Seatbelt Detection and Usage Recognition for Driver  Monitoring Systems}
\author{
    Feng HU, PhD\\
    fengh@nvidia.com\\
    Autonomous Vehicle Software, NVIDIA
}
\affiliations{

}

\usepackage{bibentry}

\pdfoutput=1

\begin{document}

\maketitle

\begin{abstract}
 Wearing a seatbelt appropriately while driving can reduce serious crash-related injuries or deaths by about half. However, current seatbelt reminder system has multiple shortcomings, such as can be easily fooled by a “Seatbelt Warning Stopper”, and cannot recognize incorrect usages for example seating in front of a buckled seatbelt or wearing a seatbelt under the arm.   General seatbelt usage recognition  has many challenges, to name a few, lacking of color information in Infrared (IR) cameras, strong distortion caused by  wide Field of View (FoV) fisheye lens, low contrast between belt and its background, occlusions caused by hands or hair, and imaging blurry. In this paper, we introduce a novel general seatbelt detection and usage recognition framework to resolve the above challenges. Our method consists of three components: a local predictor,  a global assembler, and a shape modeling process. Our approach can be applied to the driver in the Driver Monitoring System (DMS) or general passengers in the Occupant Monitoring System (OMS) for various camera modalities.  Experiment results on both DMS and OMS are provided to demonstrate the accuracy and robustness of the proposed approach.
\end{abstract}

\section{Introduction}

Wearing a seatbelt appropriately while driving can reduce serious crash-related injuries and deaths by about half, and in 2016 alone seatbelts have saved 15,000 lives, making it the greatest protection for passengers especially when combined with air bags \cite{policyImpact2021} \cite{seatbeltFacts2021}. 

 Current seatbelt reminder system has several shortcomings. First, it  can be easily fooled by a “Seatbelt Warning Stopper” \cite{seatbeltWarningStopper2021}, a device made with a small metal that fits into the buckle as if a real seatbelt is fastened. Second, incorrect seatbelt usage, such as seating in front of a buckled seatbelt or wearing a seatbelt under the arm, which is also very harmful  when crash happens \cite{intas2010seat}, cannot be detected. Third, unnecessary false alarm can happen if pets or bags are put on the passenger seats. Fourth, Current reminder system is  not enforced for the back passengers, and parents get distracted while driving for frequently checking whether their children have unbuckled up their seatbelts, e.g. for picking up a toy.

\begin{figure*}[t]
\begin{subfigure}{0.33\textwidth}
  \centering
  \includegraphics[width=0.95\linewidth]{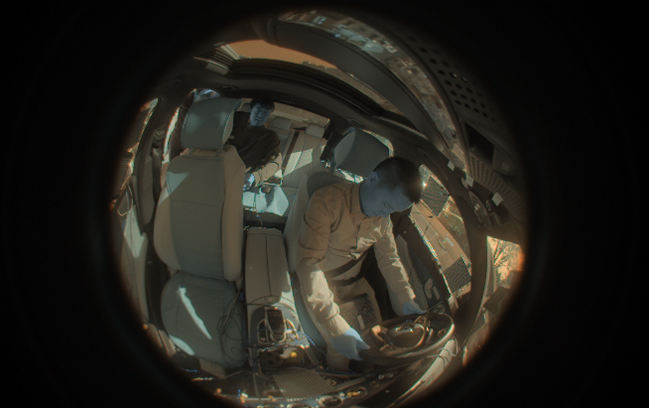}
  \caption{Color fisheye seatbelt image sample}
  \label{fig:sfig1}
\end{subfigure}
\begin{subfigure}{0.33\textwidth}
  \centering
  \includegraphics[width=0.95\linewidth]{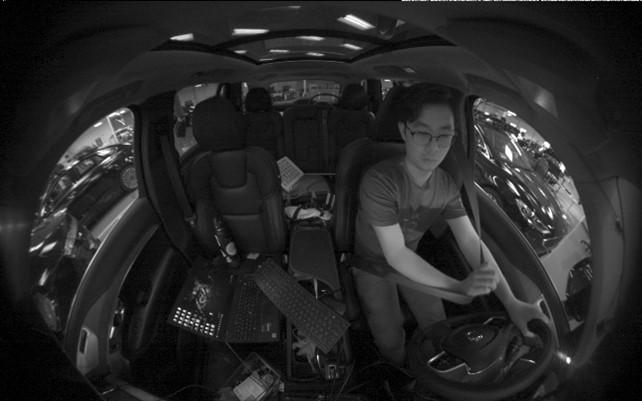}
  \caption{Gray wide FoV seatbelt image sample}
  \label{fig:sfig2}
\end{subfigure}
\begin{subfigure}{0.33\textwidth}
  \centering
  \includegraphics[width=0.95\linewidth]{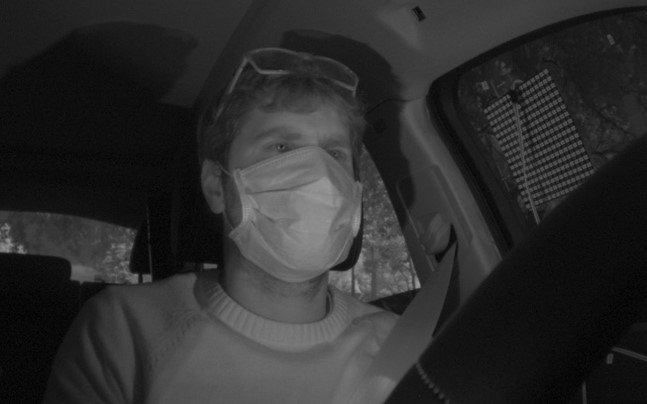}
  \caption{Gray regular FoV seatbelt image sample}
  \label{fig:sfig3}
\end{subfigure}

\caption{Sample input images for  seatbelt detection and usage recognition}
\label{fig:fig1}
\end{figure*}

Accurate and robust general seatbelt detecting and usage recognizing  is a hard problem due to many challenges.

\begin{itemize}

\item Seatbelt images may have color, e.g. Fig. \ref{fig:sfig1}, or be gray, e.g. Fig. \ref{fig:sfig2}; 
\item Super-wide Field of View (FoV) fisheye lens may be used, within which the belt is a curve as shown in Fig. \ref{fig:sfig1}, while for regular FoV lens, the belt is more like a straight line, as shown in Fig. \ref{fig:sfig3}; 
\item	The cloth and seatbelt's color may be similar, resulting in low contrast image, e.g. Fig. \ref{fig:sfig2}; 
\item	Partial of the seatbelt might be occluded by objects, e.g. hands in Fig. \ref{fig:sfig1}; 
\item	Seatbelt can be stretched into to an irregular shape, an extreme case shown in Fig. \ref{fig:sfig2}; 
\item	Image may be blurry due to object movement or vehicle vibration, e.g. Fig. \ref{fig:sfig2}.

\end{itemize}

In this paper, we propose a novel seatbelt detection and usage recognition framework to solve the above-mentioned challenges. There are three components within the framework, whose brief descriptions are shown below. 

\begin{itemize}
\item[I.]	\textbf{Local predictor}:
Scanning through the input image and predicting for each local patch window whether it is part of a seatbelt
\item[II.]	\textbf{Global assembler}:
Assembling the positive seatbelt  candidates in step $I$ by removing false positive as much as possible utilizing various attributes, such as shape, location, etc.
\item[III.]	\textbf{Shape modeling}:
Building the geometric seatbelt shape model utilizing the filtered candidates from step $II$, and mapping it on top of the original image 
\end{itemize}

We summarize our contributions as  follows. (1)	 A new  seatbelt detection and usage recognition framework that can robustly detect one or more seatbelts for both regular FoV or fisheye images. (2)	 A new local seatbelt patch predictor that  provides binary classification  for any small patch window. (3)	 A  global assembling algorithm that can effectively remove false positive seatbelt patch candidates. (4)	 A high order polynomial curve based seatbelt shape modeling functionality that  localizes the seatbelt even when occlusion exists.

The rest of the paper is organized as follows. First previous work on seatbelt detection and related topics are reviewed. Then we explain our methodology and technical details. After that we present our experiment results on various contexts. Finally we conclude our paper and provide possible future work.

\section{Related Work}

According to WHO \cite{world2009global}, more than 1.2 million people die yearly in vehicles accidents, but they would have 40\% - 50\% probability of survival if they used seatbelts properly. In addition, seatbelt can reduces serious crash-related injuries and deaths by about half via reducing the force of secondary impacts with interior strike hazards, saving more than 10K people per year in US only \cite{seatbeltFacts2021}. Building the capacity of detecting seatbelt and classifying its usage  is not only an interesting research topic, but also widely pursued by auto industrial especially by the brands that are sensitive to safety.  

Multiple researchers focus on detecting whether the drivers are wearing seatbelts or not via external traffic monitoring system for law enforcement purpose \cite{qin2014efficient} \cite{hosam2020deep}\cite{yang2020study}\cite{artan2015passenger}, where one or multiple cameras are installed on the road infrastructure such as near traffic lights. Such systems mainly focus on the seatbelt status of the driver, but not all the occupants inside the vehicle due to the limitation of the camera-vehicle relationship. In this paper, we are mainly interested in seatbelt detection and usage recognition where cameras are installed inside a vehicle. 

Fisheye or omnidirectional cameras are widely used for applications where wide Field of View (FoV) representation of the surrounding scenes are needed\cite{hu2014mobile}\cite{hu2017real}\cite{hu2018computer}. Some researchers use salient gradient feature \cite{zhou2017learning}, line-detection \cite{guo2011image} or convolutional neural network (CNN) \cite{zhou2017seat} \cite{chun2019nads}\cite{kashevnik2020seat} for regular Field of View (Fov) camera based seatbelt detection. These methods, however, cannot extend to Occupant Monitoring System (OMS) where fisheye or super wide Field of View lens are  used and images are extremely distorted. The image of a straight line in the  physical world, such as an edge of the belt is no longer straight in a fisheye image, thus simple edge or line detection based approaches fail.

In order to recognize the seatbelt usage beyond just detecting the seatbelt is on or off, such as whether the seatbelt is fastened over the shoulder or under the arm,  precise seatbelt curve locations are needed. Curve fitting with limited  point candidates is an extensively studied topic \cite{levy1959complex}\cite{guest2012numerical}. In our approach,  we adopt a polynomial curve based fitting method for fast detection. Since the input fitting data are usually noisy with outliers e.g. due to false positive  candidates, RANdom SAmpling and Consensus (RANSAC) \cite{fischler1981random} algorithm is used as well for removing the outliers. 

Seatbelts can have many types, such as 2 points, 3 points, or 4 to 6 points belts (e.g. seatbelts for child seat or racing cars), even though we believe our methodology can extend to all these types,  in this paper, we only focus on the most  popular $Y$ shape 3 points seatbelts widely used in the auto industrial. For a given car model, the 3 anchor points of a seatbelt are roughly fixed, though some anchor such as the top point may have some degree of freedom. 

Camera calibration is a process to estimate a camera's intrinsic parameters (such as focal length, distortion parameters, etc.) and extrinsic parameters (such as rotation and translation) that is necessary for many vision-based solutions \cite{li2011geometric}\cite{ren2021camera}.  If a DMS or OMS camera is installed, its intrinsic and extrinsic parameters are fixed, so camera calibration such as  Zhang's method \cite{zhang2000flexible} can be used for determining the camera's position  relative to the vehicle. Once the camera-vehicle geometric relationship is determined, we can further estimate a  seatbelt anchor's location in an seatbelt image using Perspective-n-Points (PnP) algorithm \cite{moreno2007accurate}.

\section{Methodology}

We design a bottom-up framework to tell for any given input image from a DMS or OMS system, whether the seatbelt is detected or not for a specific seat, and if detected, whether it is correctly fastened, e.g. not under the arm. 

First, from the lowest level we design a process to roughly predict for any input pixel whether it is part of a seatbelt pixel set, using itself and its neighborhood pixels' attributes.  Then, we filter out the false positive candidates and only assemble the qualified ones utilizing global seatbelt features such as  geometric location, smoothness or intensity range. Finally, we utilize the seatbelt anchors' location and the candidates survived in previous step to model the shape of the seatbelt, which is further visualized by overlapping the shape on top of the original image. 
We summarize the three steps in our framework as the local predictor, the globally assembler and the shape modeling. 

In this paper, we will use a single belt within one seatbelt image as an example, though  our proposed approach also applies even if there are multiple belts within a image. 

\subsection{Local predictor}

The local predictor predicts whether a pixel belongs to a seatbelt or it is a background pixel  using the pixel itself and its neighbors' information.  

The predictor outputs 1 or 0  representing seatbelt or non-seatbelt respectively. Due to noise, this binary classification  may have four  possible cases as shown below.

\begin{itemize}
\item 	
True Positive (TP): a pixel is part of a seatbelt, and the predictor returns 1;
\item 		
False Positive (FP): a pixel is not part of a seatbelt, but the predictor returns 1;
\item 		
False Negative (FN): a pixel is part of a seatbelt, but the predictor returns 0; and
\item 	
True Negative (TN): a pixel is not part of a seatbelt, and the predictor returns 0.
\end{itemize}

In this step, we adjust the predicting criteria to  reduce the false negative (case FN) as much as possible, even if at the cost of increasing false positive (case FP). In other word, high recall is preferred over high precision. 

\subsubsection{Notations}

\begin{figure} [t]
    \centering
    \includegraphics[width=\columnwidth]{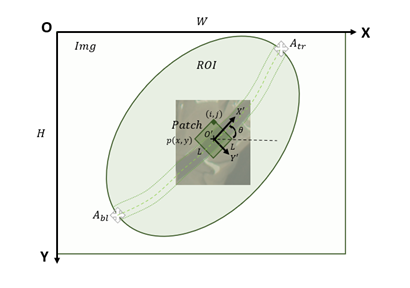}
    \caption{Seatbelt patch geometry example}
    \label{fig4:geometry}
\end{figure}

Denote $p(x,y)$ an arbitrary pixel locates at $y_{th}$ row and $x_{th}$ column within an image $Img$ as shown in Fig. \ref{fig4:geometry}. The image   width is $W$, height $H$ , and it has $K$ channels, where $K$ = 3 (or 4) for color images, and $K$ = 1 for gray images.

 We define a square  around $p(x,y)$ as its neighborhood set. Assume the length of the square is $L= 2k+1$ pixels, where  $k$ = 1,2,3…,  the set $S$ can be formulated as: 

\begin{equation}\label{equ2:setSDefinition2}
S=\{p_{ij} |  0 \leq |i - y| \leq k,0 \leq |j - x| \leq k\}           \end{equation}                        

Since seatbelt can extend along different directions in an image, as shown in Fig. \ref{fig:sfig1} and Fig. \ref{fig:sfig2}, isotropic predicting, i.e., checking along different directional angle for each pixel, is necessary. The local predictor returns binary prediction result vector $\bar{r}(x,y)$ for each pixel as defined in Equ. \ref{equ3:rDefinition}.

\begin{equation}\label{equ3:rDefinition}
\bar{r}(x,y) = (r_{\theta_0},r_{\theta_1}, \ldots, r_{\theta_D})^T    \end{equation}   
where $D$ is the total number of directions, $\theta_0$, $\theta_1$, $\ldots$, $\theta_D$ are evenly spaced angles within the range of $[0,\pi)$. 

Define
\begin{equation}\label{equ4:thetaDDefinition}
r_{\theta_d} = 
\begin{cases}
1,  if f_{criteria} (x,y,\theta_d ) \geq T_{threshold} \\
0,  otherwise\\
\end{cases}
\end{equation}
where $d = 0, 1, \ldots, D - 1$. The criteria function $f_{criteria} (x, y, \theta_d )$ determines for any pixel $p(x,y)$ within image $Img$ and a specific direction $\theta_d$, the patch image generated using $p$ and its neighborhood $S$ at this direction is a seatbelt component or not. We will describe with more details about the criteria function later in this section.

In addition we define the local predictor $f_{predictor}(x, y)$ as a scale function of $\bar{r}(x,y)$:
\begin{equation}\label{equ5}
    f_{predictor}(x, y) = \sum_{d=0}^{D-1} w_{\theta_d} \cdot r_{\theta_d}
\end{equation}
where $w(\theta_d),  d = 0, 1, \ldots, D-1$ are weights assigned to different directions.

\subsubsection{Observations}
Assume we have fetched a seatbelt patch for an arbitrary pixel together with its $L\times L$ neighbors in a targeted direction (patch direction), as shown in Fig. \ref{fig4:geometry}, we  observe multiple properties.

\begin{figure*}[t]
\begin{subfigure}{0.5\columnwidth}
  \centering
  \includegraphics[width=0.7\columnwidth]{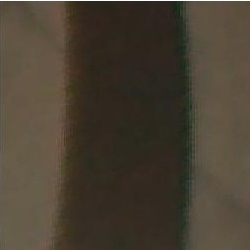}
  \caption{}
  \label{fig3:a}
\end{subfigure}
\begin{subfigure}{0.5\columnwidth}
  \centering
  \includegraphics[width=0.7\columnwidth]{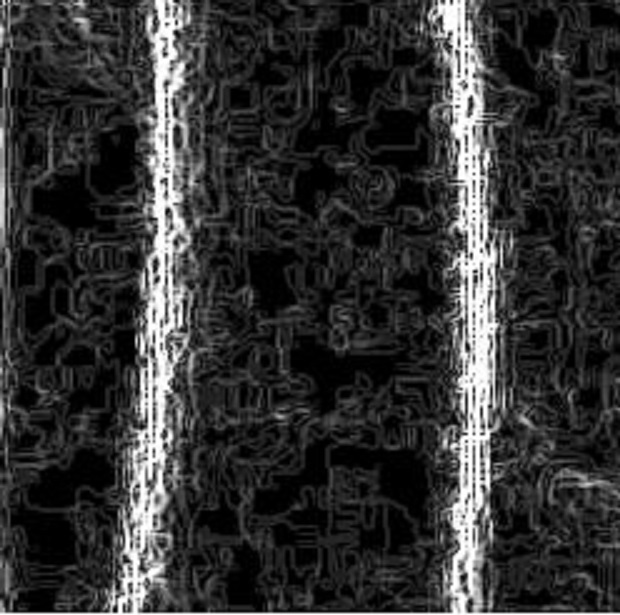}
  \caption{ }
  \label{fig3:b}
\end{subfigure}
\begin{subfigure}{0.5\columnwidth}
  \centering
  \includegraphics[width=0.7\columnwidth]{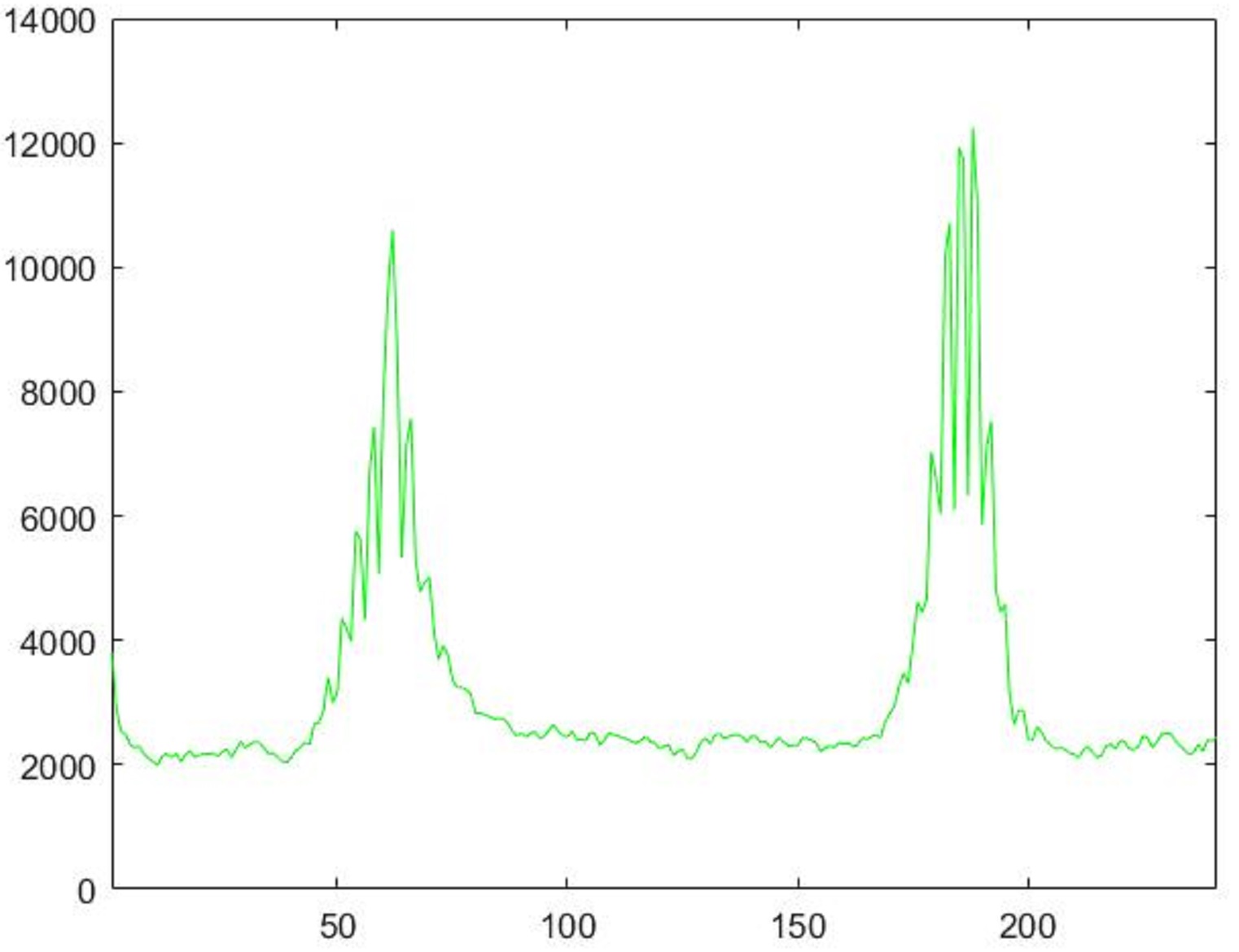}
  \caption{}
  \label{fig3:c}
\end{subfigure}
\begin{subfigure}{0.5\columnwidth}
  \centering
  \includegraphics[width=0.7\columnwidth]{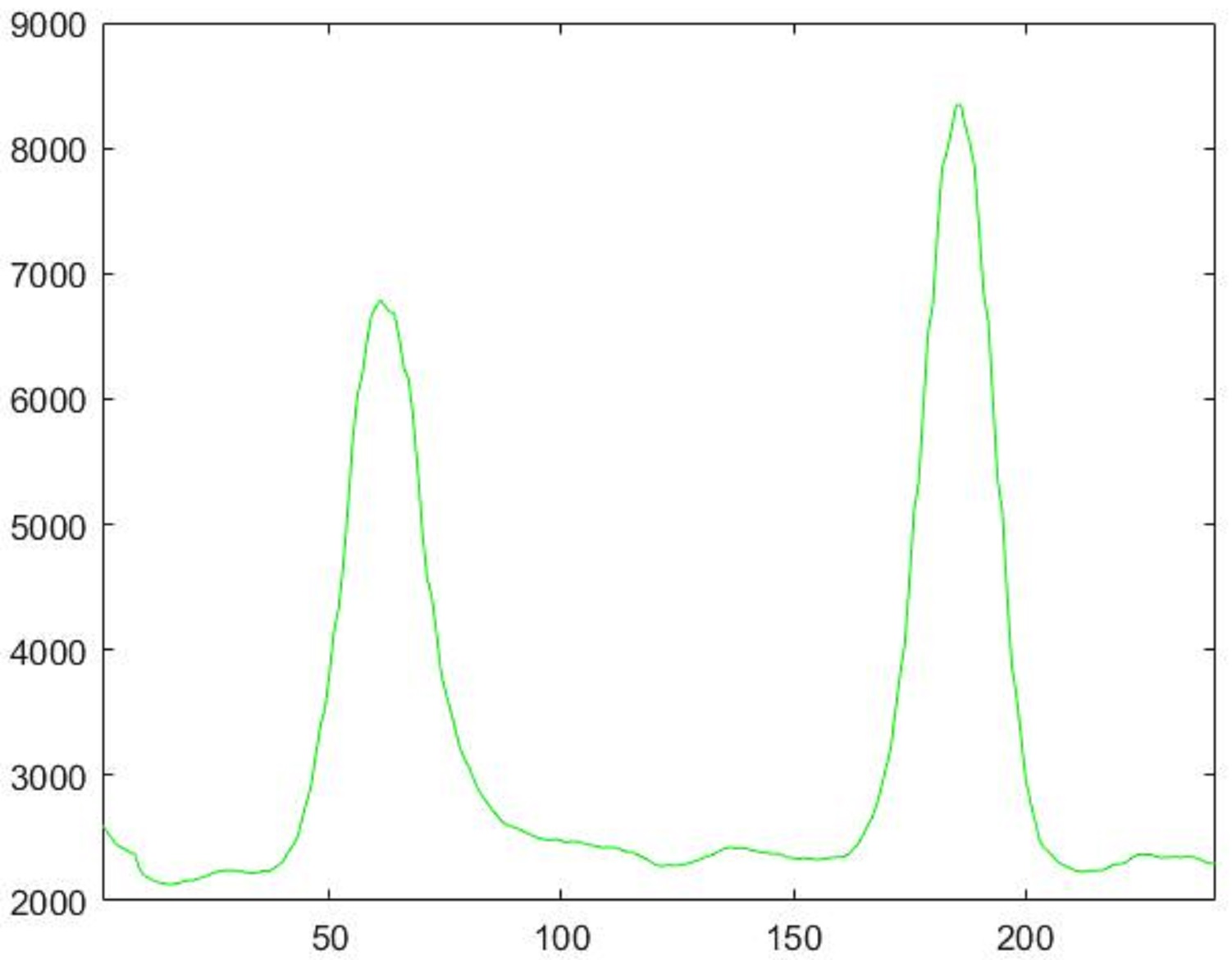}
  \caption{ }
  \label{fig3:d}
\end{subfigure}\\
\begin{subfigure}{0.5\columnwidth}
  \centering
  \includegraphics[width=0.7\columnwidth]{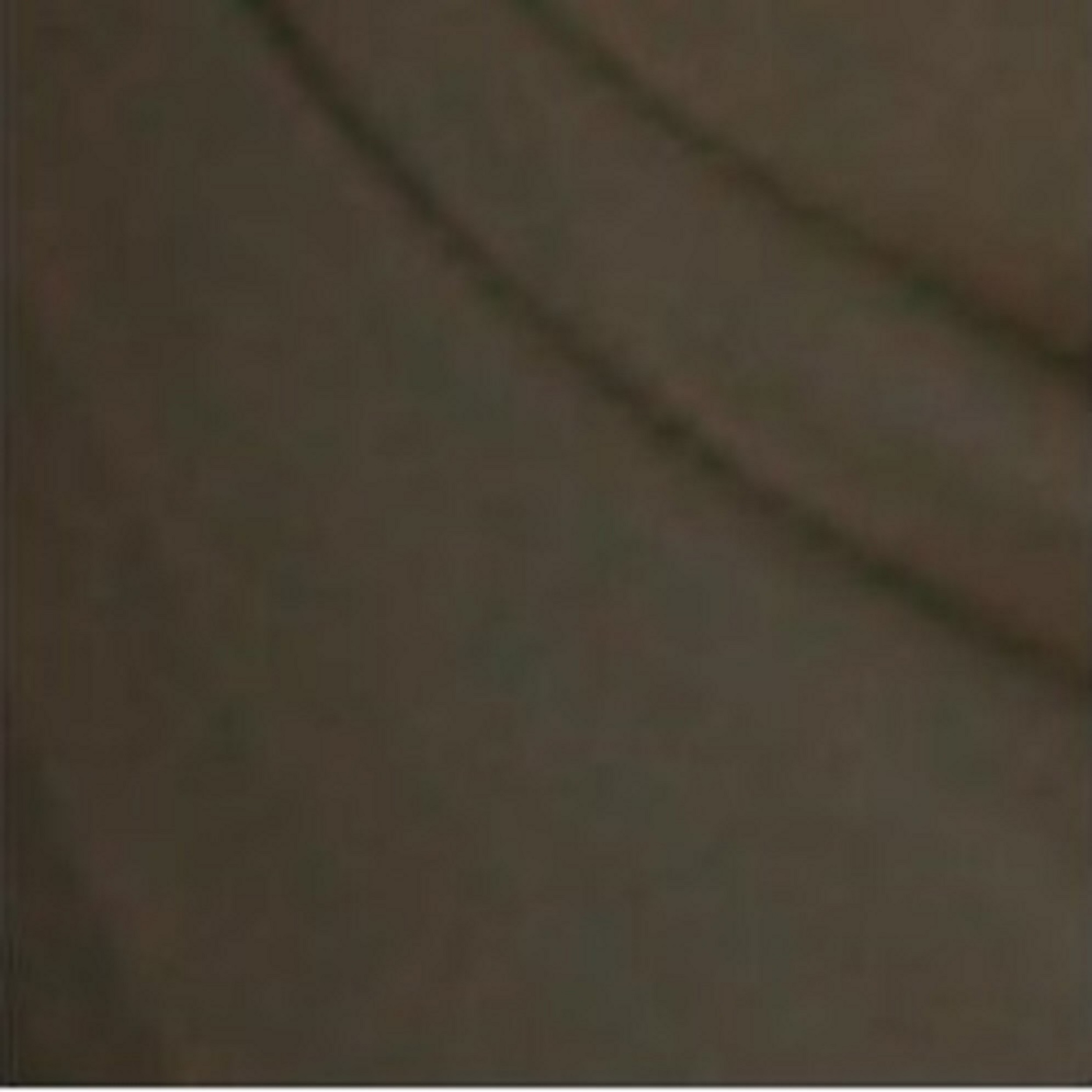}
  \caption{ }
  \label{fig3:e}
\end{subfigure}
\begin{subfigure}{0.5\columnwidth}
  \centering
  \includegraphics[width=0.7\columnwidth]{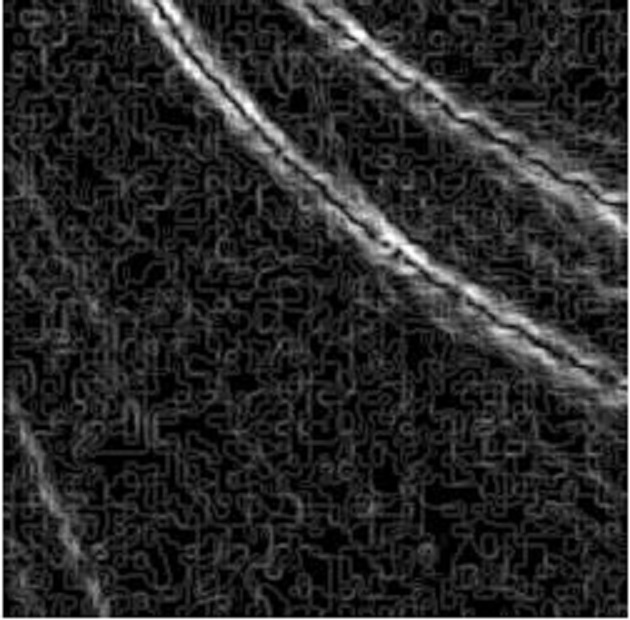}
  \caption{}
  \label{fig3:f}
\end{subfigure}
\begin{subfigure}{0.5\columnwidth}
  \centering
  \includegraphics[width=0.7\columnwidth]{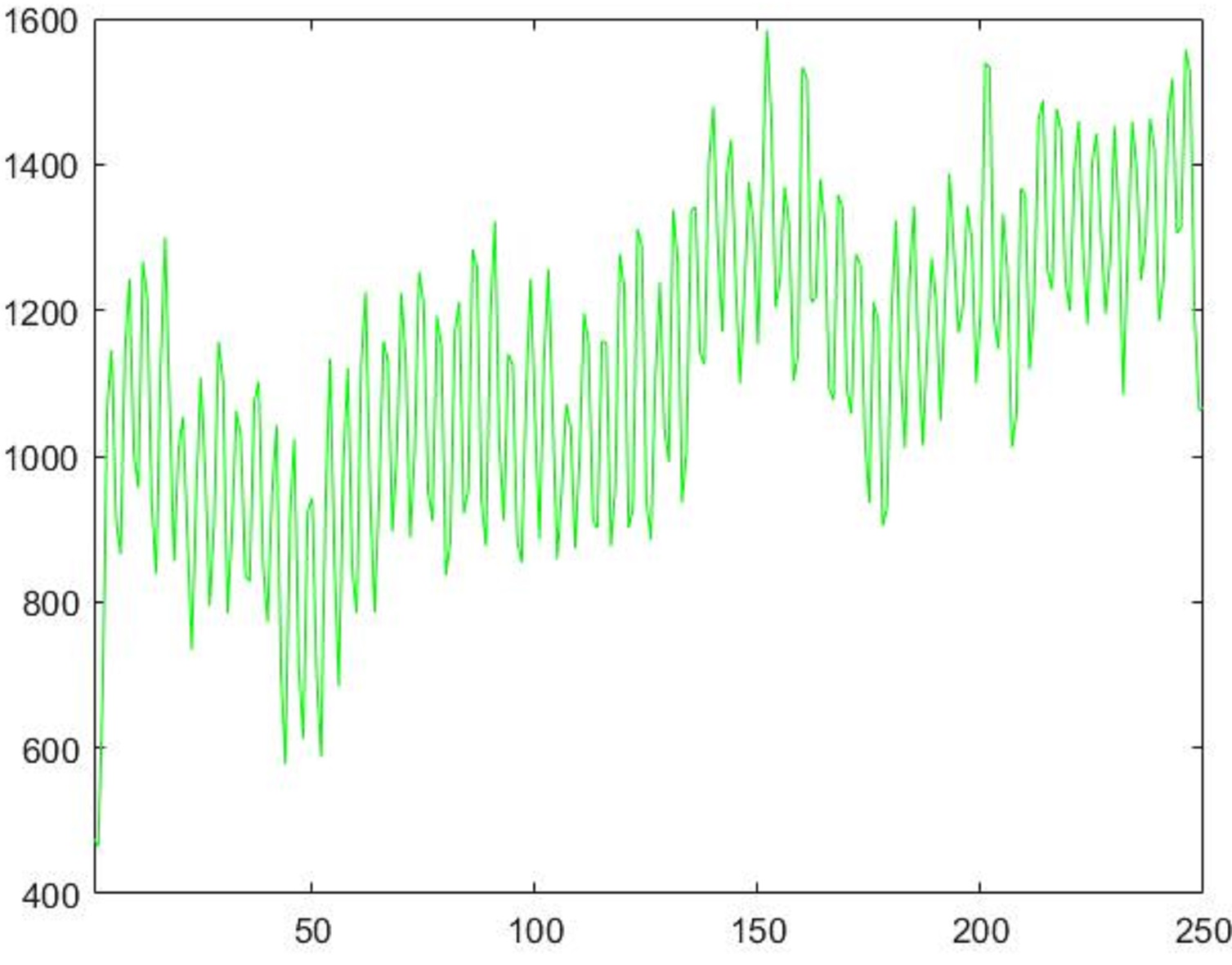}
  \caption{}
  \label{fig3:g}
\end{subfigure}
\begin{subfigure}{0.5\columnwidth}
  \centering
  \includegraphics[width=0.7\columnwidth]{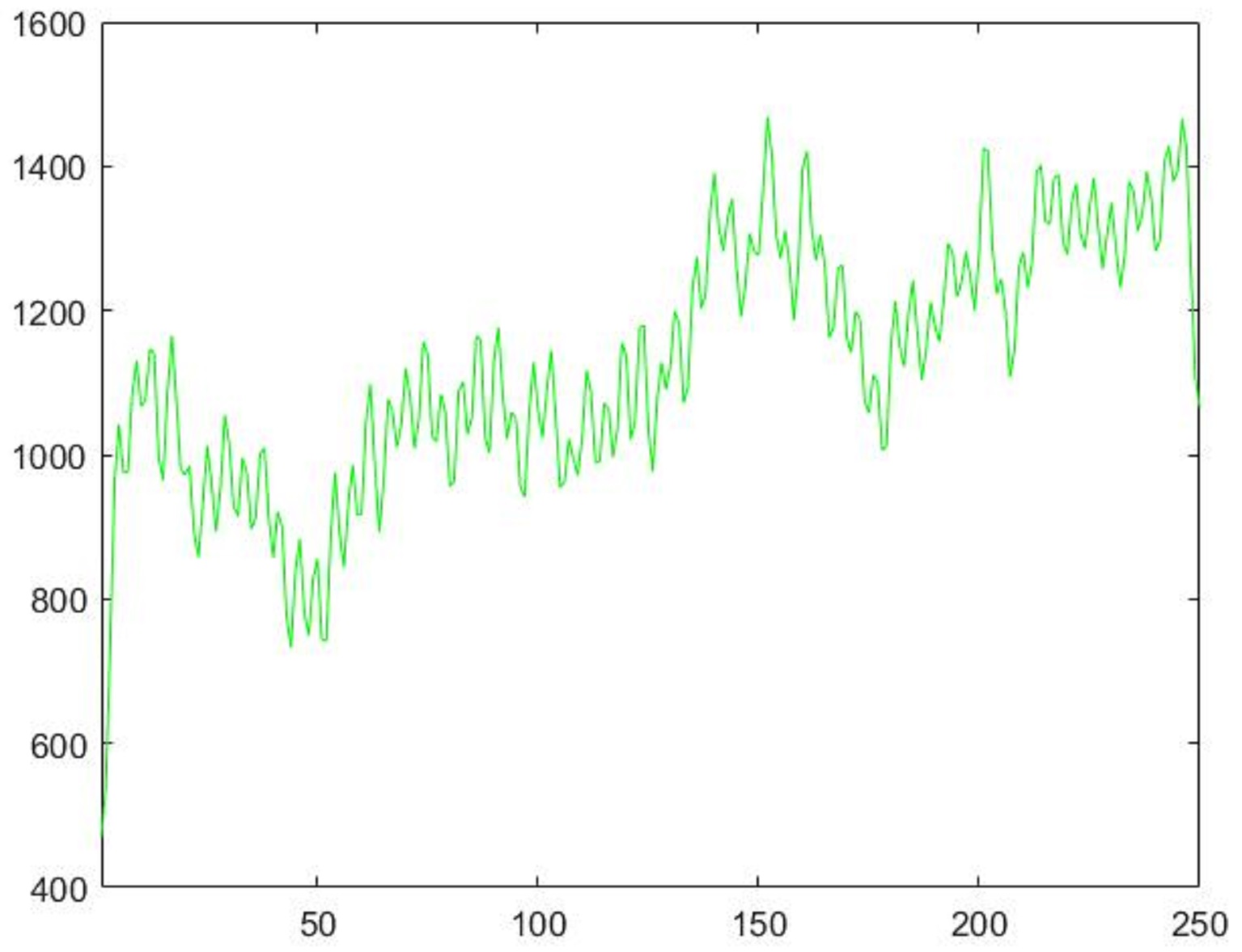}
  \caption{}
  \label{fig3:h}
\end{subfigure}
\caption{Seatbelt and non-seatbelt patches. (a) Seatbelt patch sample; (b) Augmented seatbelt patch; (c) Seatbelt patch projected curve; (d) Smoothed seatbelt curve; (e) Non-seatbelt patch sample; (f) Augmented non-seatbelt patch; (g) Non-seatbelt patch projected curve; (h) Smoothed non-seatbelt curve.}
\label{fig:fig3SamplesWith8Figs}
\end{figure*}

\begin{itemize}
\item 
Observation (I): structured edges. Seatbelt edges generate two usually parallel lines in the patch direction even in fisheye images, if the patch is small enough.

\item 
Observation (II): intensity or saturation range. Seatbelt usually comes with black, grey, or tan color. Though pixel values change in some context such as illumination changes, we still can exclude extreme values experimentally.

\item 
Observation (III): surface smoothness. Standard seatbelts usually have similar texture along the belts thus the pixels within seatbelt region will have limited variety in terms of smoothness.
\end{itemize}

Denote $f_{structure}(x,y,\theta_i )$, $f_{intensity}(x,y,\theta_i )$, and $f_{smoothness}(x,y,\theta_i )$ as binary functions modeling above three observations, then:

\begin{multline}\label{equ6}
    f_{criteria} (x,y,\theta_i ) = f_{structure} (x,y,\theta_i )  \cap \\ f_{intensity} (x,y,\theta_i )  \cap f_{smoothness} (x,y,\theta_i )
\end{multline}

\subsubsection{Criteria (I): structure}

We will first discuss how to identify and localize candidate seatbelt edges, then we will talk about how to use it to distinguish seatbelt pixel from non-seatbelt noise. 

Each seatbelt patch, i.e. a candidate pixel plus its neighbors in a square, has a patch direction by definition, and for each patch, we only check whether there is a seatbelt parallel structure along the patch direction.

As shown in Fig. \ref{fig4:geometry}, for an arbitrary patch centered at pixel $p(x,y)$ with patch direction $\theta$, we  retrieve its $L * L$ pixel values from image $Img$ using Equ. \ref{equ7}.

\begin{equation}\label{equ7}
    \begin{bmatrix}
        x' \\ y'
    \end{bmatrix} = 
    \begin{bmatrix}
        cos \theta & -sin \theta \\ 
        sin \theta & cons \theta
    \end{bmatrix} (  
    \begin{bmatrix}
        j \\ i
    \end{bmatrix} -
    \begin{bmatrix}
        \frac{1}{2} L \\ \frac{1}{2} L
    \end{bmatrix}    
    )    +
    \begin{bmatrix}
        x \\ y
    \end{bmatrix}    
\end{equation}
where $i, j$ = 1,2,\ldots,$L$.

Fig. \ref{fig3:a} and Fig. \ref{fig3:e} show two sample seatbelt and non-seatbelt patches, where the seatbelt patch has a parallel structure, i.e. two parallel vertical lines, along patch direction but the non-seatbelt patch does not. If we localize the positions of the structure, we have detected the seatbelt boundary within this patch.  

We utilize 2-dimensional gradient operation to capture the magnitude of sudden intensity change, and accumulate along patch direction to obtain the patch 1D curve representation. Denote $f_{patch}$ $(x,y,\theta,j)$ as the resulting curve representation after this projection.

\begin{equation}\label{equ8}
    f_{patch}= \sum_{i=1}^{L} \sqrt{(\frac{\partial f (x,y,\theta,i, j)}{\partial j} )^2 + (\frac{\partial f(x,y,\theta,i,j)}{\partial i} )^2}
\end{equation}
where $j$ is the variable indexing patch columns, $i$ indexing rows, and $f(x,y,\theta,i,j)$ is patch intensity function. Fig. \ref{fig3:b} and Fig. \ref{fig3:f} show the patch images after gradient operations, and the correspondent  curve representations are shown in Fig. \ref{fig3:c} and Fig. \ref{fig3:g}.

We can further smooth the curves e.g. using  Savtzsky-Golay algorithm \cite{gorry1990general} to get a smoothed curve  $f_{patch}' (x,y,\theta,j)$ as shown in Fig. \ref{fig3:d} and Fig. \ref{fig3:h}.

\begin{figure}
    \centering
    \includegraphics[width=0.6\columnwidth]{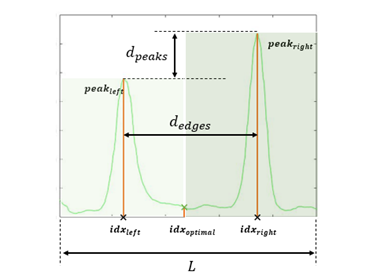}
    \caption{Structured edges localization}
    \label{fig5:edge}
\end{figure}

Observe that a seatbelt patch with structured edges will have two peaks as shown in Fig. \ref{fig5:edge}, while non-seatbelt patch will have random curves. We characterize this observation with two variables: the inter-edge distance $d_{edges}$ and peak value difference $d_{peaks}$.  $f_{structure} (x,y,\theta_i )$ is therefore defined in Equ. \ref{equ10}.

\begin{multline}\label{equ10}
f_{structure} = 
\begin{cases}
1,  if \tau_{min} \leq d_{edges} \leq  \tau_{max} \cap \\ \rho_{min} \leq \frac{d_{edges}}{max(peak_{left}, peak_{right}) } \leq \rho_{max}\\
0,  otherwise
\end{cases}
\end{multline}
where $\tau_{min}$, $\tau_{max}$, $\rho_{min}$ and $\rho_{max}$ are experimental thresholds.

To calculate $d_{edges}$ and $d_{peaks}$, we need to identify the two peak values ( $peak_{left}$ and $peak_{right}$), and their respective indexes ($idx_{left}$ and $idx_{right}$) for a given curve.

The problem is converted to a one-dimensional two-classes classification problem via finding the optimal cutting position $idx_{optimal}$ among all the possible $idx$, where $idx = 1, 2, \ldots, L$, by maximizing the inter-class distance. 

\begin{multline}\label{equ11}
    idx_{optimal} = arg \max_{1 \leq idx \leq L} w_{left} * (\mu_{left} - \mu_L )^2 + \\ w_{right} * (\mu_{right} - \mu_L )^2 
\end{multline}
where, \[w_{left}= \sum_{j=1}^{idx} f_{patch}^{'} (x,y,\theta,j)\] 
 
\[w_{right}= \sum_{j=idx + 1}^{L} f_{patch}^{'} (x,y,\theta,j)\] 

 \[\mu_{left}= \sum_{j=1}^{idx} j * f_{patch}^{'} (x,y,\theta,j)\] 
 \[\mu_{right}= \sum_{j=idx + 1}^{L} j * f_{patch}^{'} (x,y,\theta,j)\] 
   \[\mu_{L}= \sum_{j=1}^{L} j * f_{patch}^{'} (x,y,\theta,j)\]

After we  figure out $idx_{optimal}$, $idx_{left}$ and $idx_{right}$ are calculated by searching for the maximum elements position in $[1,idx_{optimal}]$ and $(idx_{optimal},L]$. Their respective function values are therefore  $peak_{left}$  and $ peak_{right}$. 

\subsubsection{Criteria (II): intensity}
Seatbelt pixel's intensity varies per seatbelt models, and may change when  environmental illumination changes. However,   in major Driver Monitoring Systems (DMS), we use active InfraRed (IR) to illuminate and spectral filter to remove external illumination's influence. So for a specific working seatbelt, its intensity distribution has patterns and can be learned from experiments, which we can utilize to assist the local predicting. 

Assume $\delta_{min}$ and $\delta_{max}$ are the lower and upper intensity bounds, and denote weighted intensity for $f_{patch} (x,y,\theta)$ as $d_{intensity}$. 

\begin{equation}\label{equ12}
    d_{intensity}=\sum_{i = 1}^{L} \sum_{j = 1}^{L} w_{ij} * f_{patch} (x,y,\theta,i,j) 
\end{equation}
where  \[w_{ij}=\frac{1}{2\pi \sigma^2 } e^{-\frac{(\frac{L}{2}-i)^2+(\frac{L}{2}-j)^2}{2\sigma^2}}\] are Gaussian distributed weights assigned to each pixel in the seatbelt patch. Finally we have the definition of $f_{intensity}$ in Equ. \ref{equ13}.

\begin{equation}\label{equ13}
f_{intensity} = 
\begin{cases}
1,  if \delta_{min}  \leq d_{intensity}  \leq \delta_{max}\\
0,  otherwise\\
\end{cases}
\end{equation}

\subsubsection{Criteria (III): smoothness}
The surface of a seatbelt is usually smooth without abrupt intensity changes, while the background are much noisy. Define the intensity variance within a Region of Interest (ROI) $\omega$ as $d_{smoothness}$.

\begin{equation}\label{equ14}
    d_{smoothness}=\sum_{i = x - \omega}^{x + \omega} \sum_{j = y - \omega}^{y + \omega} ( f_{patch} (i, j) - f_{patch} (x, y))^2
\end{equation}

We therefore have the definition of $f_{smoothness}$ in Equ. \ref{equ15}, where $\varphi_{min}$ and $\varphi_{max}$ are experimental thresholds.

\begin{equation}\label{equ15}
f_{smoothness} = 
\begin{cases}
1,  if \varphi_{min}  \leq d_{smoothness}  \leq \varphi_{max}\\
0,  otherwise\\
\end{cases}
\end{equation}

\subsection{Global assembler}

The local predictor generates large quantity of seatbelt pixel candidates, including true positive and false positive results.  In this section, we will present how to use global assembler to remove the inaccurate candidates.

Since the seatbelt is always tightened by the built-in springs no matter fastened or not, the location distribution of seatbelt pixels observed via a fixed camera can be roughly determined by the three anchors.

In the following, we will introduce how to utilize this prior knowledge to globally filtering false positive candidates. Furthermore, we will also discuss how to automatic calculate anchors’ locations for a specific camera-vehicle configuration.

\subsubsection{Global mask generation}

Once a DMS or OMS camera's pose is configured, i.e. both camera location and its orientation  are determined, the seatbelt's footprint in the image will be within a limited area, which we denoted as global mask for this configuration. We will first introduce how to leverage camera calibration and vehicle model to automatically generate the seatbelt global location mask.

Denote the camera is calibrated with intrinsic parameters $K$, and extrinsic parameters $R$ and $T$, also the seatbelt has top right anchor  $A_{tr} (X_{tr},Y_{tr},Z_{tr})$ and bottom left anchor $A_{bl} (X_{bl},Y_{bl},Z_{bl})$ inside the vehicle 3D model. Then the anchors’ correspondent coordinates in the image $A_a (x_a,y_a ),a=tr,bl$ can be calculated with Equ. \ref{equ16}.

\begin{equation}\label{equ16}
    \begin{bmatrix}
        x_a \\ y_a \\ 1
    \end{bmatrix} = s * K *
    \begin{bmatrix}
        R & T \\ 
        0 & 1
    \end{bmatrix}  
    \begin{bmatrix}
        X_a \\ Y_a \\ Z_a \\ 1
    \end{bmatrix}    
\end{equation}
where $s$ is a scale factor, and $a=tr,bl$.

According to our experience, majority of location distribution of the seatbelt will fall into an ellipse using the two anchors as ends of the long axis. Denote the major axis distance $d_{major}$ as shown in Equ. \ref{equ17}.

\begin{equation}\label{equ17}
    d_{major}=  \sqrt{(x_{tr}-x_{bl} )^2 + (y_{tr}-y_{bl})^2}
\end{equation}

Meanwhile, the minor axis distance can be obtained via analyzing people’s seatbelt fastening habits statistically. Assume the minor axis distance is $d_{minor}$, we have the definition of the global mask set $S_{location}$ shown in Equ. \ref{equ18}. 

\begin{equation}\label{equ18}
S_{location}=\{(x_{local},y_{local}) | \frac{x_{local}^2}{(\frac{1}{2}d_{major})^2} + \frac{y_{local}^2}{(\frac{1}{2}d_{minor})^2} \leq 1\}          
\end{equation}

Note that the $(x_{local}, y_{local})$ are defined in ellipse local coordinate system, where the origin $O_{elipse}$ is at the middle point of segment $A_{bl}$ $A_{tr}$ with image coordinate $(\frac{x_{tr}+x_{bl}}{2},\frac{y_{tr}+y_{bl}}{2})$,   x-axis pointing from $A_{bl}$ to $A_{tr}$, $y$-axis obtained via clockwise rotating $x$-axis for 90 degrees. Assume the correspondent coordinate of $(x_{local}, y_{local})$  in image coordinate is $(x_{global},y_{global})$. 

\begin{equation}\label{equ19}
    \begin{bmatrix}
        x_{global} \\ y_{global}
    \end{bmatrix} = 
    \begin{bmatrix}
        cos \theta_s & -sin \theta_s \\ 
        sin \theta_s & cos \theta_s
    \end{bmatrix}  
    \begin{bmatrix}
        x_{local} \\ y_{local}
    \end{bmatrix}    +
    \begin{bmatrix}
        \frac{x_{tr} + x_{bl}}{2} \\  \frac{y_{tr} + y_{bl}}{2}
    \end{bmatrix}     
\end{equation}
where \[\theta_s = acos \frac{x_{tr} - x_{bl}}{\sqrt{(x_{tr}-x_{bl})^2 + (y_{tr}-y_{bl})^2}}\]

With Equation \ref{equ18} and \ref{equ19}, we can obtain the location mask $S_{location}$ in the image coordinate. The seatbelt pixel candidates proposed by local predictors outside of the global mask will be filtered out.

\subsection{Shape modeling}

After locally predicting and globally assembling, we obtain, for each seatbelt region of interest, a set of pixel candidates. However, the expected output is a continuous seatbelt curve from anchor to anchor, with which we derive the usage category. In this subsection, we will explain how to  model the seatbelt shape using the survived pixel candidates. 

Assume  the set of candidates from local predictor $S_{predictor}$ and the location mask $S_{location}$, we define the candidates set for shape modeling $S_{modeling}$ in Equ. \ref{equ22}.

\begin{multline}\label{equ22}
S_{modeling}=\{(x, y) | (x, y) \in S_{location} \cap \\ f_{prediction}(x, y) \leq \gamma_{pre} \}          
\end{multline} 
where $\gamma_{pre}$ is a threshold for local predicting. 

We model the seatbelt shape using high order polynomial curve. However, the seatbelt may be stretched away while driving, an example of which is shown in Fig. \ref{fig:sfig2}. So curve fitting directly in the image coordinate system is not feasible, since for some points in the $x$-axis, there are multiple correspondent $y$ values. We solve this problem by first fitting the curve in the local ellipse coordinate system and then convert the result back into the image coordinate system using Equ. \ref{equ19}.

Assume the observed points are $(x_i,y_i)$, we model the curve using Equ. \ref{equ24}. 

\begin{equation}\label{equ24}
y_i= \beta_0+ \beta_1 x_i^1+ \beta_2 x_i^2+ \ldots + \beta_N x_i^N   
\end{equation} 
where $(\beta_0,\beta_1,\beta_2 \ldots \beta_N )$ are coefficients and $N$ is the polynomial order, e.g. $N = 4$. 

Since there are still outliers in the shape modeling stage,  M-estimator sample consensus (MSAC) algorithm \cite{torr2000mlesac} can be used to remove the outliers.

\section{Experiments}

We conduct our   experiments for data collected by both Driver Monitoring System (DMS) using regular Field of View (FoV) InfraRed (IR) camera and Occupancy Monitoring System (OMS) using super wide FoV fisheye IR camera. 

\begin{figure*}[t]
\begin{subfigure}{\columnwidth}
  \centering
  \includegraphics[width=0.95\columnwidth]{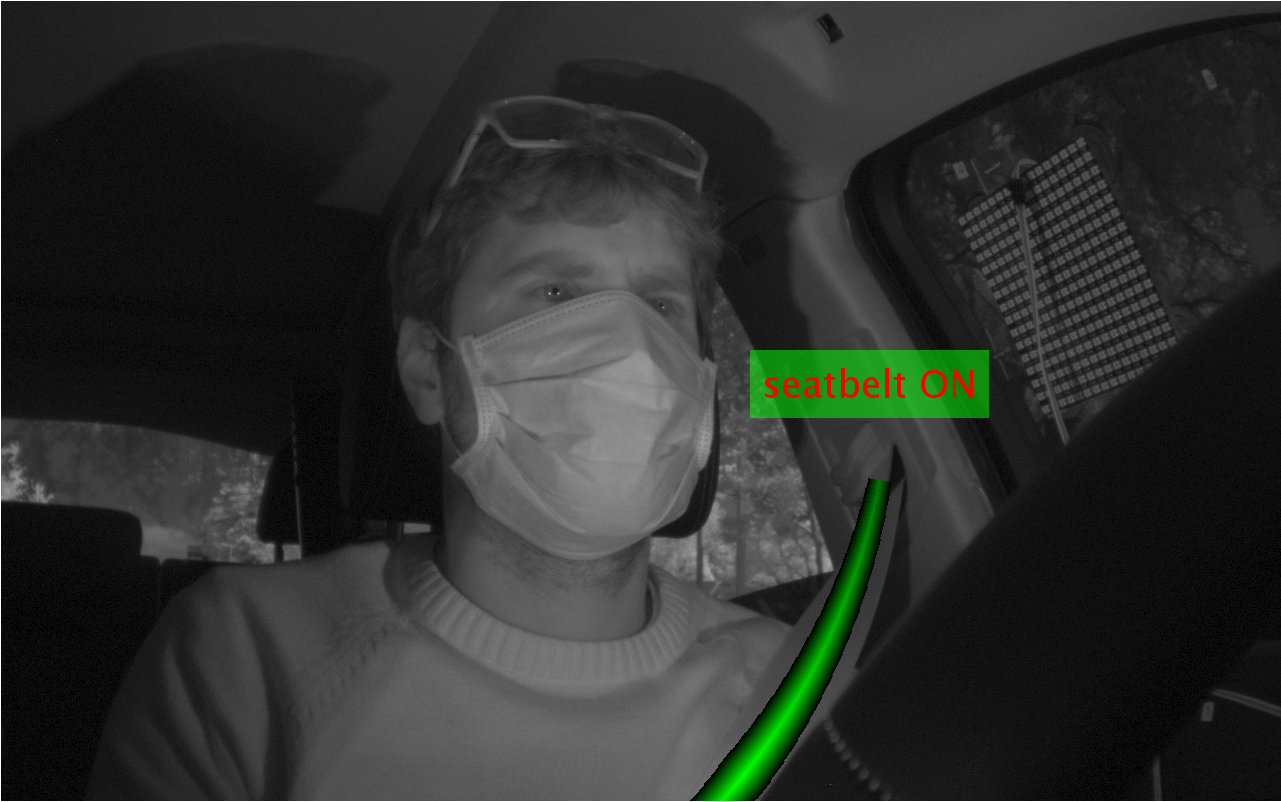}
  \caption{Regular camera: case on}
  \label{fig:dms_on}
\end{subfigure}%
\begin{subfigure}{\columnwidth}
  \centering
  \includegraphics[width=0.95\columnwidth]{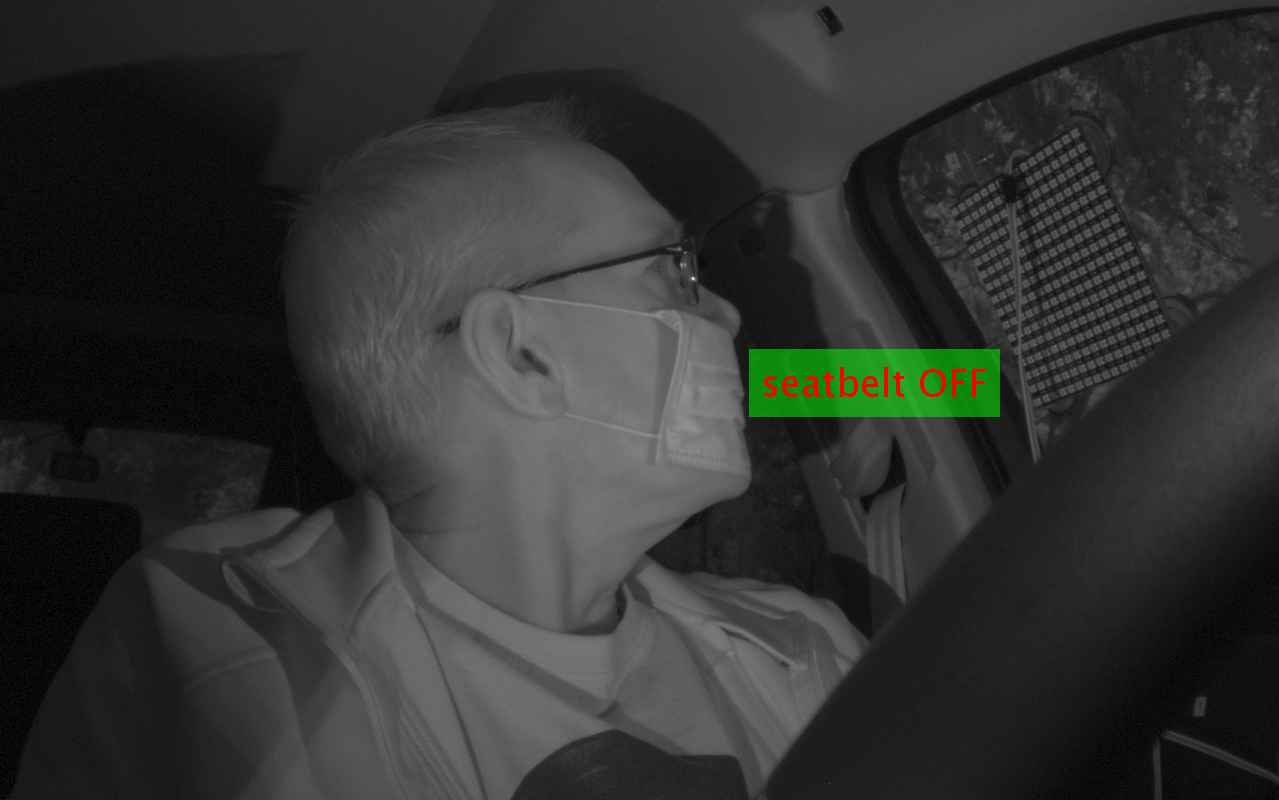}
  \caption{Regular camera: case off}
  \label{fig:dms_off}
\end{subfigure}
\begin{subfigure}{\columnwidth}
  \centering
  \includegraphics[width=0.95\columnwidth]{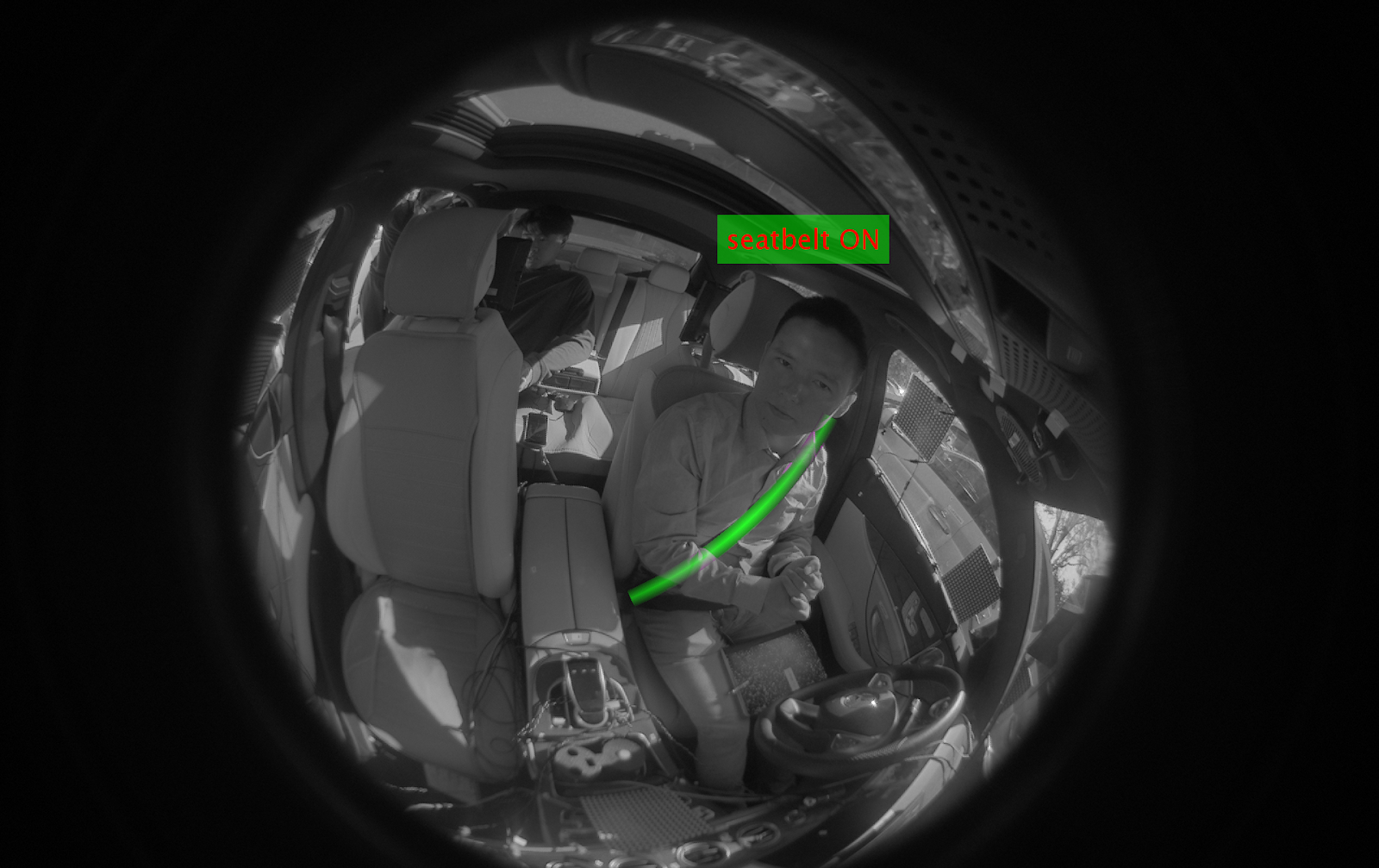}
  \caption{Fisheye camera: case on}
  \label{fig:oms_on}
\end{subfigure}%
\begin{subfigure}{\columnwidth}
  \centering
  \includegraphics[width=0.95\columnwidth]{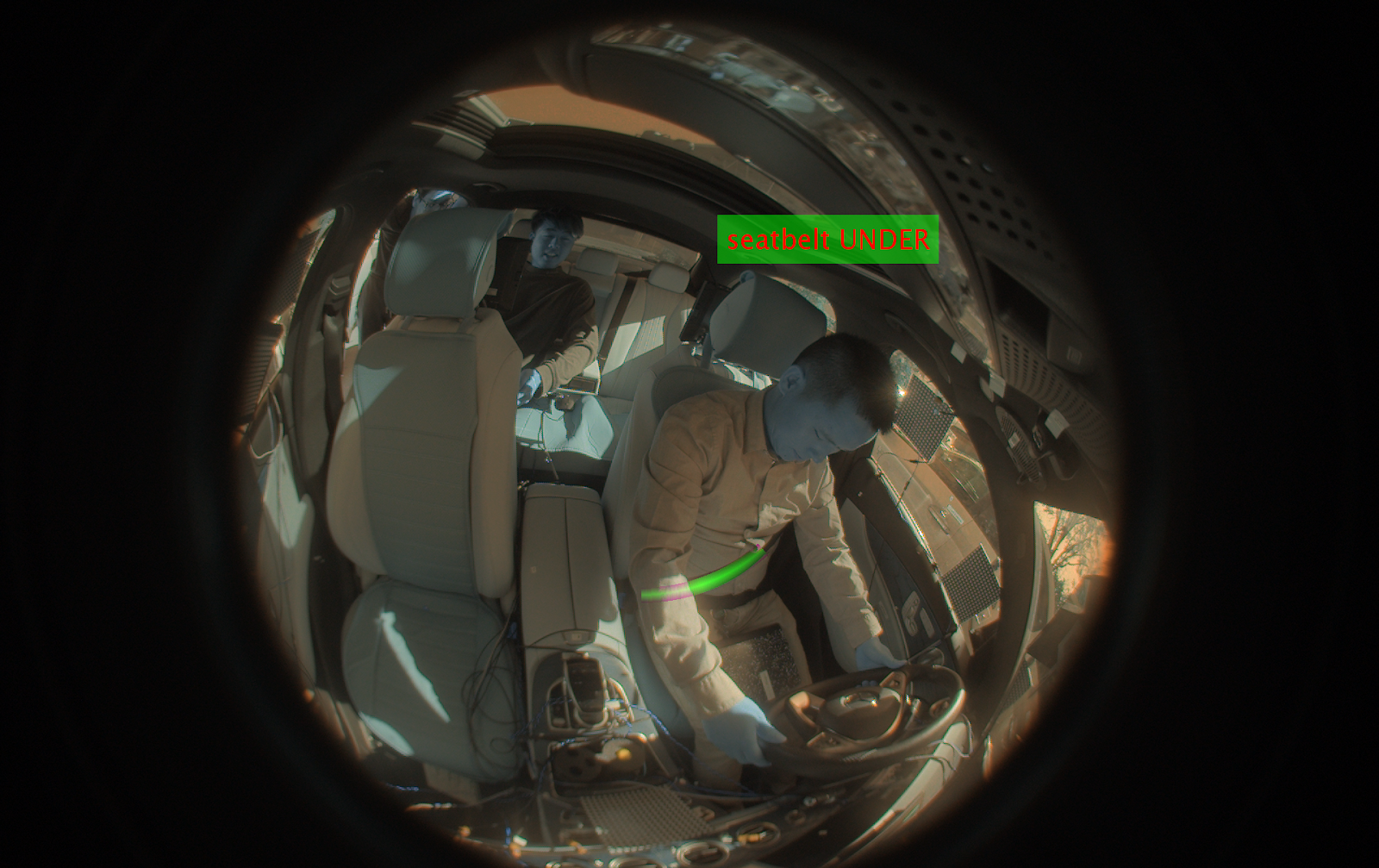}
  \caption{Fisheye camera: case under}
  \label{fig:oms_under}
\end{subfigure}
\caption{Seatbelt detection and recognition for regular and fisheye camera}
\label{fig:OMS}
\end{figure*}

\subsection{Usage categorization}
We define three seatbelt usage category as shown below.
\begin{itemize}
\item 	
seatbelt ON: seatbelt is correctly fastened above a user's shoulder
\item 		
seatbelt UNDER: seatbelt is fastened in a incorrect way, i.e. under a user's arm
\item 		
seatbelt OFF: no seatbelt is observed within the image
\end{itemize}

Note that for the seatbelt off case, it can be caused by seating in front of a fastened seatbelt, a "Seatbelt Warning Stopper" is used while the actual seatbelt is off, or the user is forgotten to fasten the seatbelt.  There are ways to distinguish from these three sub-cases,  e.g. reading the buckle sensor information from the vehicle's Controller Area Network (CAN) signals. However, it is beyond this paper's scope and we in any case  send out a warning message when no seatbelt is observed.

To distinguish between "seatbelt ON" and "seatbelt UNDER", we calculate the angles between image $x$ axis and the line segment of left bottom anchor and points in the seatbelt curve. If a seatbelt is fastened under the arm, the angle is much smaller compared with correctly fastened over the shoulder.

\subsection{DMS results}
Fig. \ref{fig:dms_on} shows seatbelt detection and recognition results for Driver Monitoring System using regular Field of View camera, where the modeled seatbelt  is rendered using green curves and its category printed above the curve in red fonts.  Fig. \ref{fig:dms_off} shows a seatbelt detection result when the seatbelt is not fastened by a different subject. 

We test our algorithm with 407 images with seatbelt on and 974 images with seatbelt off from multiple users. The $dms.mp4$ (\url{https://bit.ly/2ZWddlW}) shows  sample results.

The recognized seatbelt usage results can  be encoded and sent out to the vehicle system, e.g. using CAN, and proper protocol can be defined to reminder the users for correction, such as using seatbelt warning symbols in the dashboard, seatbelt warning sound, or voice reminder.

\subsection{OMS results}
Fig. \ref{fig:oms_on} shows a sample result where seatbelt is correctly fasten in fisheye image in Occupant Monitoring System. 
Fig. \ref{fig:oms_under} shows another sample image where seatbelt is incorrectly fasten via under the arm. 

We test our algorithm with 319 images with seatbelt ON, 312 with seatbelt OFF and 318 with seatbelt UNDER. We also ask the subject to move around to test the robustness of the algorithm.

The $oms.mp4$ (\url{https://bit.ly/3k8dARJ}) shows  results with various usage category on full video sequences. The method correctly recognizes 97.49\% images for seatbelt ON, 100\% when seatbelt is off, and 98.11\% when seatbelt is fastened under the arm.

Note that seatbelt may be occluded, e.g. by the hand of the user in many frames. However, since our approach is not only just detecting seatbelt pixels, but also modeling the entire curve using the seatbelt candidates, it can recover the occluded part and output a complete seatbelt.

\subsection{Extreme case}

\begin{figure} 
    \centering
    \includegraphics[width=\columnwidth]{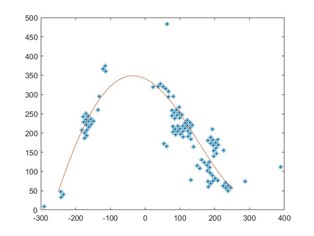}
    \caption{Sample candidates and curve fitting results (red line)}
    \label{fig6:shape_modeling}
\end{figure}

\begin{figure} 
    \centering
    \includegraphics[width=0.7\columnwidth]{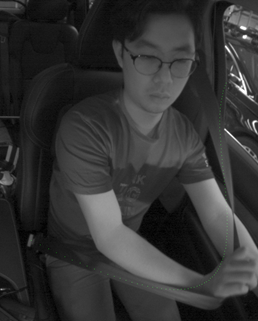}
    \caption{seatbelt result for extreme case where belt is stretched away}
    \label{fig:away}
\end{figure}

For extreme cases where the seatbelt is stretched away, as shown in Fig.\ref{fig:sfig2}, it is very challenging for regular line  detection based seatbelt  algorithm to work. Fig. \ref{fig6:shape_modeling} shows some intermediate results of our algorithm. The blue points are the assembled seatbelt candidates  and the red line is the  modeled shape in the ellipse coordinate system. Fig. \ref{fig:away} shows the zoomed in correspondent result after mapping the curve in Fig. \ref{fig6:shape_modeling} to the original image using green dots.

\section{Conclusion}
In this paper, a novel  seatbelt detection and usage recognition method is proposed  with three components: local predictor, global assembler, and shape modeling. The approach utilizes a hybrid of  image color/intensity, shape, smoothness, and location to effectively localize a seatbelt from noise background. It works with both regular FOV or fisheye images, color or IR images, and for different categories (such as seatbelt ON, OFF, UNDER, or AWAY). Experiment results are shown to demonstrate  its effectiveness and robustness under various conditions.

In the future, we will extend our experiments to more vehicle models and more camera-vehicle configurations.

\bibliography{aaai22}


\end{document}